\documentclass[manuscript]{acmart}
\AtBeginDocument{%
  \providecommand\BibTeX{{%
    \normalfont B\kern-0.5em{\scshape i\kern-0.25em b}\kern-0.8em\TeX}}}
    
\renewcommand\footnotetextcopyrightpermission[1]{} % removes footnote with conference information in first column

\setcopyright{acmcopyright}
\copyrightyear{2023}
\acmYear{2023}
\acmDOI{10.XXXX/XXXXXXX.XXXXXXX}

\acmConference[CHI 2023]{The ACM Conference on Human Factors in Computing Systems}{April 23-28, 2023}{Hamburg, Germany}

\usepackage[nolist]{acronym}
\usepackage{subcaption}
\usepackage{xcolor}
\usepackage{booktabs}
\begin{document}

\title[This Was (Not) Intended]{This Was (Not) Intended: How Intent Communication and Biometrics Can Enhance Social Interactions With Robots}

\author{Khaled Kassem}
\affiliation{%
  \institution{TU Wien}
  \country{Austria}}
\email{khaled.k.kassem[at]tuwien.ac.at}

% , 
\author{Alia Saad}
\affiliation{%
  \institution{University of Duisburg-Essen}
  \country{Germany}}
\email{alia.saad[at]uni-due.de}

\renewcommand{\shortauthors}{Kassem et al.}

\let \oldcite \cite
\renewcommand{\cite}[1]{~\oldcite{#1}}

\begin{abstract}
Socially Assistive Robots (SARs) are robots that are designed to replicate the role of a caregiver, coach, or teacher, providing emotional, cognitive, and social cues to support a specific group. SARs are becoming increasingly prevalent, especially in elderly care. Effective communication, both explicit and implicit, is a critical aspect of human-robot interaction involving SARs. Intent communication is necessary for SARs to engage in effective communication with humans. Biometrics can provide crucial information about a person's identity or emotions. By linking these biometric signals to the communication of intent, SARs can gain a profound understanding of their users and tailor their interactions accordingly. The development of reliable and robust biometric sensing and analysis systems is critical to the success of SARs.
In this work, we focus on four different aspects to evaluate the communication of intent involving SARs, existing works, and our outlook on future works and applications. 
\end{abstract}

\begin{acronym}%[\hspace{3cm}]
\acro{hci}[HCI]{Human-Computer Interaction}
\acro{hri}[HRI]{Human-Robot Interaction}
\acro{hrc}[HRC]{Human-Robot Collaboration}
\acro{ar}[AR]{Augmented Reality}
\acro{vr}[VR]{Virtual Reality}
\acro{xr}[XR]{Mixed Reality}
\acro{ems}[EMS]{Electrical Muscle Stimulation}
\acro{ros}[ROS]{Robot Operating System}
\acro{ct} [CT] {Collaborative Task}
\acro{eeg} [EEG] {electroencephalogram}
\end{acronym}

\newcommand{\alia}[1]{\textsf{\textcolor{red}{\textbf{Alia:} \textit{#1}}}}
\newcommand{\kk}[1]{\textcolor{blue}{{\textbf{KK:} \textit{#1}}}}

\maketitle

\section{Introduction}

Socially Assistive Robots (SARs) can be deployed in situations where they can replicate the role of a caregiver or coach. SARs can support, motivate, and provide emotional support without physical contact, to encourage learning, development, or therapy~\cite{mataric2016socially}. Previous findings showed the promising results of SARs in education and elderly care~\cite{Robinson2022}.

% explicit communication
For assistive robots to be accepted as trustworthy and effective, they must be able to engage in effective communication with humans, who may be individuals receiving support from SARs, or human "peers" in a similar position of care, e.g. a nurse or teacher. Effective communication is crucial in human-robot interaction (HRI) involving SARs, as SARs should both communicate their intent and comprehend human intent, especially in a collaborative setting~\cite{KalpagamGanesan2018}.

%implicit communication
In addition to explicit communication, implicit communication has a vital role in social interaction~\cite{Matari2017}. Biometrics such as facial expressions, body posture, and tone of voice are also used by humans for implicit communication, and can continuously provide crucial information about a person's emotional state, personality traits, and level of engagement. By using these biometric signals as implicit modalities for intent communication, SARs can gain a more complete understanding of their users and tailor their interactions accordingly. Moreover, SARs being able to emulate human non-verbal cues in social situations makes interaction smoother\cite{mataric2016socially}. As such, the development of reliable and robust biometric sensing and analysis systems is critical to the success of SARs. 

%tie a bow around it
In this paper, we reflect on the role of implicit and explicit bidirectional intent communication in establishing connections and fostering trust in SARs. Additionally, we look at how findings from different contexts (e.g. industry) can be applied to the more social settings of SARs. We focus on four aspects of HRI with SARs, namely: \textbf{transparency}, \textbf{trust}, \textbf{error-recovery}, and \textbf{adaptation}.
\begin{comment}
We summarize the research directions in form of questions as follows:

\begin{itemize}
    \item How can socially assistive robots adapt to communicate transparently and recover from errors during interaction?
    \item In what ways can socially assistive robots be adapted to better serve diverse user populations?
\end{itemize}

    \item \alia{How should SARs transparently communicate when reasoning is based on complex processes?} 
    \item \alia{How do SARs recover from unintended/incorrect actions through communication with the user?}  
    \item \alia{How should SARs approaches be adapted for specific populations?} % multiple specific groups -> multiple social interaction needs   
\end{comment}

\section{Transparency}\label{transparency}
% definition
% what was done? on which type of robots? limitations?
% foresights 
%\alia{I think we could start with: 
The first criterion in evaluating the intent of communication for SARs is transparency. %}
Previous research highlights the importance of effective communication skills for robots, which can enhance their reliability, predictability, and transparency to humans. As a result, users are more likely to trust and accept the technology\cite{Chadalavada2015}. By extension, transparency is critical for building trust and acceptance of SARs. In this section, we explore the role of transparency in enhancing the acceptance of SARs and how intent communication can help in achieving transparency. We also discuss the potential applications of behavioral biometrics in promoting transparency in interactions with SARs.

%Effective communication between humans and robots is essential for smooth interaction and successful teaming. SARs collaborating with humans should use the same clear and two-way communication used in human teams\alia{dont like this statement, what do you mean}. To ensure this, there needs to be a shared language and channel for an effective exchange of information. This requires a clear expression of states and goals, which helps in building transparency between humans and SARs. As noted by \citet{Shively2017}, designers must ensure that SARs (as a form of automation) can communicate clearly and present information in a way that fits human mental models.

Similar to traditional human interactions, clear two-way communication is crucial for successful interactions between humans and robots. Consequently, communication should be mutually understandable by both humans and robots for an effective exchange of information. With a clear expression of states and goals, transparency could be achieved. As noted by \citet{Shively2017}, designers must ensure that SARs (as a form of automation) can communicate clearly and present information in a way that fits human mental models.

Explicit communication in HRI can be visual, auditory, or haptic~\cite{chareview2018,Kumar2021}. Visual and auditory methods are the most studied methods in HRI~\cite{chareview2018} and are the most suited for communication at a distance, such as in a situation involving SARs. Visual indicators used in Robot-to-Human communication include lights and projection, while auditory ones include speech as well as noises such as beeps. However, the context of interaction controls the type of communication. For example, SAR supporting care in a hospital, where noises or beeps could disturb patients, should consider other forms of indicating its intent than auditory solutions.

Robots can also signal their intent to humans implicitly, For example, \citet{Reinhardt2021} developed a non-verbal cue for robots in a situation of a robot and a human walking into a "bottleneck", where the robot moving backward indicates the intent to yield way to pedestrians, calling it a "back-off" movement.  Study participants reported that this action effectively conveyed the intention of yielding priority to pedestrians and improved the efficiency of their interaction with the robot. 

Meanwhile, intent communication in the other direction (i.e. \emph{human-to-Robot}) can also be non-verbal, utilizing gestures and motions that are natural in human-human communication. Previous work showed that a human's gaze can be an indication of interest or attention~\cite{irfan2020using}. Additionally, biometric sensors such as eye trackers in the context of socially assistive robotics can be used to gauge human attention~\cite{Mutlu2012}. Sensors such as eye trackers and thermal cameras are now available commercially and can be used to infer a variety of emotional states or responses ~\cite{Abdrabou2017,Kassem2017,Salah2018}.

We can envision more examples where SARs utilize implicit intent communication. For example, an embodied SAR can utilize anthropomorphic features such as limbs or eyes to convey information non-verbally. Moreover, an embodied SAR can use its own gaze to exhibit interest and guide the human's attention toward particular objects or areas. However, excessive eye contact could give unsociable impressions. Thus, it is crucial to ensure that implicit communication utilized by SARs is still clear to the human, without being disturbing or intimidating. As a result, there is a need to investigate how SARs can mimic implicit cues used by humans in social interaction while maintaining transparency and clarity. 

\textbf{We recognize the potential of utilizing and recognizing implicit cues such as gaze, body language, and gestures in SAR design, improving the smoothness of interactions with humans.}

To name another example: a SAR in a teaching role could gauge the level of engagement based on human students' gaze. A human avoiding eye contact during conversation can be a sign of discomfort. However, with implicit communication comes the challenge of clarity. The challenge for the SAR is to understand subtle differences which would indicate different human behaviors or emotional states, and intuitively process them the way a human would.  

\textbf{We argue for the use of lower-cost sensors to build future SARs can make the technology more accessible and affordable.}
\section{Trust}
%\cite{Hald2021} "explanations alone are not sufficient to increase human-computer trust after robot mistakes."

%\cite{Liu2021} "robot autonomy leads to a range of negative perceptions in humans. After watching videos of autonomous robots, people rated them as more difficult to control, more intelligent, less desirable, less user intention, and eerier."

%\cite{Aroyo2017} People will trust a robot to help if they ask it for help. However, they are reluctant to ask for help in the first place until they really need it.

%\cite{Esterwood2022} robot personality appears to have a significant and positive impact on acceptance.

Trust and acceptance are essential components of interaction with SARs. Transparent and clear explanations (for behavior and decisions) alone are insufficient to increase user trust after robots make mistakes\cite{Hald2021}. Instead, building trust involves more than simply communicating intentions. For example, caregiving SARs are expected to act with some degree of autonomy, as a human caregiver would. However, robot autonomy can lead to negative perceptions and a lack of trust; in a study by \citet{Liu2021}, participants rated autonomous robots as more difficult to control, less desirable, and eerier after seeing them on video. This is an example of how fostering trust in SARs requires a delicate balance between autonomy and compliance.

Moreover, using personal biometrics to authenticate users can ensure secure interactions between humans and robots. For instance, in therapeutic settings, ensuring that only authorized individuals can access confidential information such as personal medical records would enhance the sense of trust ~\cite{Kellmeyer2018}. Similarly, monitoring biometrics such as heart rate variability, skin conductance, and facial expressions can improve human-robot interaction by providing feedback on the user's emotional state, thus allowing the robot to adjust its behavior accordingly and offer more appropriate assistance\cite{Lin2020}.

\textbf{We conclude that building trust in SARs involves more than explainability. It requires understanding the user's privacy and security concerns. Personal biometrics can be an implicit part of intent communication and can be used to enhance security and improve the user's trust in SARs.}

\section{Error Recovery}

Errors can happen in a situation involving SARs due to miscommunication, misinterpretation, or an incomplete mental model of the situation at hand, to name a few. When errors happen, attribution and recovery are important factors that can influence the human perception of trust, acceptance, and engagement with SARs \cite{Kaniarasu2014}. Previous work has shown that while people like robots that give them credit and take the blame for mistakes, explicit attribution of blame might not be advisable in a task-oriented scenario where the human must collaborate with the self-blaming robot~\cite{Kaniarasu2014}. The effectiveness of error-recovery strategies is situational, depending on the task, context, and severity of failure~\cite{Brooks2016}. Therefore, understanding the appropriate strategies for error recovery and the context-associated impact of error attribution is crucial for the success of SARs.

Error \emph{discovery} is a prerequisite to \emph{recovery}. We argue that biometrics can play a crucial role in SAR's recognition of errors by recognizing resulting human non-verbal cues such as a surprised or frustrated facial expression. Such implicit communication can help SARs dynamically adjust error recovery methods based on effectiveness. 

Additionally, SARs must recover from errors safely and without creating new errors. For example, a SAR making an error during a physical therapy session must recognize the error and correct it without harming the user.\\

\textbf{We postulate that SARs must be designed to recognize errors in real-time, assess the impact of the error, and take appropriate action to correct the error while minimizing risk to the user}.

%\kk{reflect}

\section{Personalization and Adaptation}

\emph{Recognition} comes hand-in-hand with \emph{personalization}. SARs are designed to provide assistance to individuals, and personal biometrics can help to personalize their responses and behavior to the individual's specific needs. This has led to adaptation becoming a core research direction for socially assistive robotics~\cite{mataric2016socially}. Behavioral biometric data can provide real-time feedback on the user's physical and emotional state, enabling continuous non-obtrusive adjustments. {\'A}lvarez-Aparicio used the Laser Imaging Detection and Ranging (LIDAR) sensor on a non-humanoid robot to identify individuals based on their gait~\cite{alvarez2022biometric}, reaching a precision score of 88\%. Although their approach was not specifically designed for SARs, it illustrates a possible use case where a robot identifies a user walking from a distance, understands their preferences, and acts accordingly.

While much work has been done on biometric-based identification in non-robotic applications, only a few studies have explored their use in SARs. Carcagn\`{i} et al.~\cite{carcagni2015soft,carcagni2015visual} introduced a system that allows a humanoid robot to recognize the age and gender of multiple persons in the robot's camera field of view based on their facial features. Though their approach reached results with an accuracy of 67.4\%, user feedback was not investigated. On the other hand, similar works adopted face recognition in a long-term study for a personalized SAR for cardiac rehabilitation on a single patient~\cite{irfan2020using,irfan2022personalised}. Despite the poor performance of user recognition, positive reactions were reported throughout the overall interaction. Moreover, \citet{Chan2012} utilize biosensors to recognize the affective states of users, so that the SAR can adapt accordingly. \textbf{We therefore recognize the role of implicit user responses in designing adaptable responses for such social contexts involving SARs.}

We anticipate the importance of exploring the potential usage of biometrics in improving communication between humans and SARs. Such exploration could expand beyond personalization applications, ensuring more secure and private interactions in an implicit and continuous way. Additionally, we expect adopting non-traditional biometrics to gain more attention, rather than the commonly used physiological types such as face recognition and fingerprint. However, such investigations should align with understanding the users' experience. \textbf{We argue that personalization for socially assistive robots should not solely rely on accuracy results, but should also include target demographic inputs in the form of focus groups or post-study interviews.}
\section{Future Outlook: A Hospital Scenario}
Consider a scenario in which a SAR is utilized in a pediatric hospital to provide support for children during medical procedures. We examine how the preceding sections can guide the SAR's design:
\begin{enumerate}
\item \textbf{Transparency}: The SAR can offer clear explanations of its actions to both children and their caregivers to facilitate understanding of the robot's decisions. It uses implicit cues such as gaze and tone of voice to make the interaction smoother.
\item \textbf{Trust}: Incorporating a friendly demeanor into the SAR's design can establish trust by providing words of encouragement and reassurance during procedures. To maintain patient confidentiality, biometrics are employed, and sensitive or personal information is accessible only to caregivers.
\item \textbf{Error-recovery}: The SAR recognizes negative reactions to its suggestions and adapts its behavior accordingly.
\item \textbf{Adaptation}: Monitoring biometric data, such as facial expressions, allows the SAR to tailor its behavior to better suit a child's emotional state and enhance their overall experience. The SAR utilizes age-appropriate language when communicating with younger patients, as opposed to adult caregivers.
\end{enumerate}

\section{Conclusion}

This paper examines the intersection of research in personal biometrics and intent communication, and explores how findings in these fields can be leveraged to improve the usability and acceptability of SARs. Based on our reflection on previous work, we identify several areas for future research, including \textbf{developing guidelines for transparent communication in SARs that can be applied universally, adapting implicit cues for unique situations such as those involving cultural differences, creating biometric privacy and security guidelines for SARs, incorporating commercial biosensors in SAR design for accessibility and affordability}, and \textbf{allowing target user groups to influence SAR design through, for example, focus groups}. We believe that research in these directions can produce results that enhance the acceptability of SARs.

\begin{comment}
In this paper, we reflected on prominent fields of research in HCI and HRI, and how different findings can be applied to enhance the usability and acceptability of SARs. We highlight some areas for future research work in SARs including:
\begin{itemize}
    \item Balancing implicit communication with transparency to avoid discomfort or misunderstandings for humans.
    \item Developing guidelines for transparent communication in SARs that can be universally applied.
    \item Adapting interpretation of implicit cues for unique situations, e.g., across cultural differences.
    \item Creating biometric privacy and security guidelines for SARs.
    \item Incorporating commercial biosensors in SAR design for accessibility and affordability.
    \item Allowing target user groups to influence SAR design, e.g. through focus groups.
\end{itemize}
\end{comment}

\bibliographystyle{ACM-Reference-Format}
\bibliography{references}

\end{document}